\documentclass[letterpaper, 10 pt, conference]{ieeeconf}
\IEEEoverridecommandlockouts
\usepackage{amsmath}
\usepackage{enumerate}
\usepackage{graphicx}
\usepackage{cite}
\usepackage{hyperref}

\usepackage{balance}

\graphicspath{./images}
\DeclareGraphicsExtensions{.pdf,.eps, .jpg, .png, .PNG}

\title{\LARGE \bf Multi-directional Bicycle Robot for Steel Structure Inspection}
\vspace{-15pt}
\author{Son Thanh Nguyen$^{1}$, Hai Nguyen$^{2}$, Son Tien Bui$^{3}$, Van Anh Ho$^{3}$, Hung Manh La$^{1}$, \textit{IEEE Senior Member}
\thanks{This work is supported   by the U.S. National Science Foundation (NSF) under grants NSF-CAREER: 1846513 and NSF-PFI-TT: 1919127, and the U.S. Department of Transportation, Office of the Assistant Secretary for Research and Technology (USDOT/OST-R) under Grant No. 69A3551747126 through INSPIRE University Transportation Center. 
The views, opinions, findings and conclusions reflected in this publication are solely those of the authors and do not represent the official policy or position of the NSF and USDOT/OST-R.
}
\thanks{$^{1}$The authors are with the Advanced Robotics and Automation (ARA) Lab, Department of Computer Science and Engineering, University of Nevada, Reno, NV 89557, USA. $^{2}$Northeastern University, Boston, MA 02115, USA. $^{3}$Japan Advanced Institute of Science and Tech., Japan. Corresponding author: Hung La, email: {\tt\small{hla@unr.edu}}.}
}

\begin{document}
\vspace{-15pt}

\maketitle
\begin{abstract} 

This paper presents a novel design of a multi-directional bicycle robot, which targets inspecting general ferromagnetic structures including complex-shaped structures. The locomotion concept is based on arranging two magnetic wheels in a bicycle-like configuration with two independent steering actuators. This configuration allows the robot to possess multi-directional mobility. An additional free joint helps the robot naturally adapt to non-flat and complex surfaces of steel structures. The robot has the biggest advantage to be mechanically simple with high mobility. Besides, the robot is equipped with sensing tools for structure health monitoring. We demonstrate the deployment of our robot to perform steel rust detection on steel bridges. The final inspection results are visualized as 3D models of the bridges together with marked locations of detected rusty areas.
\end{abstract}


\section{Introduction}

\begin{figure}[ht]
\centerline{\includegraphics[width=0.95\linewidth]{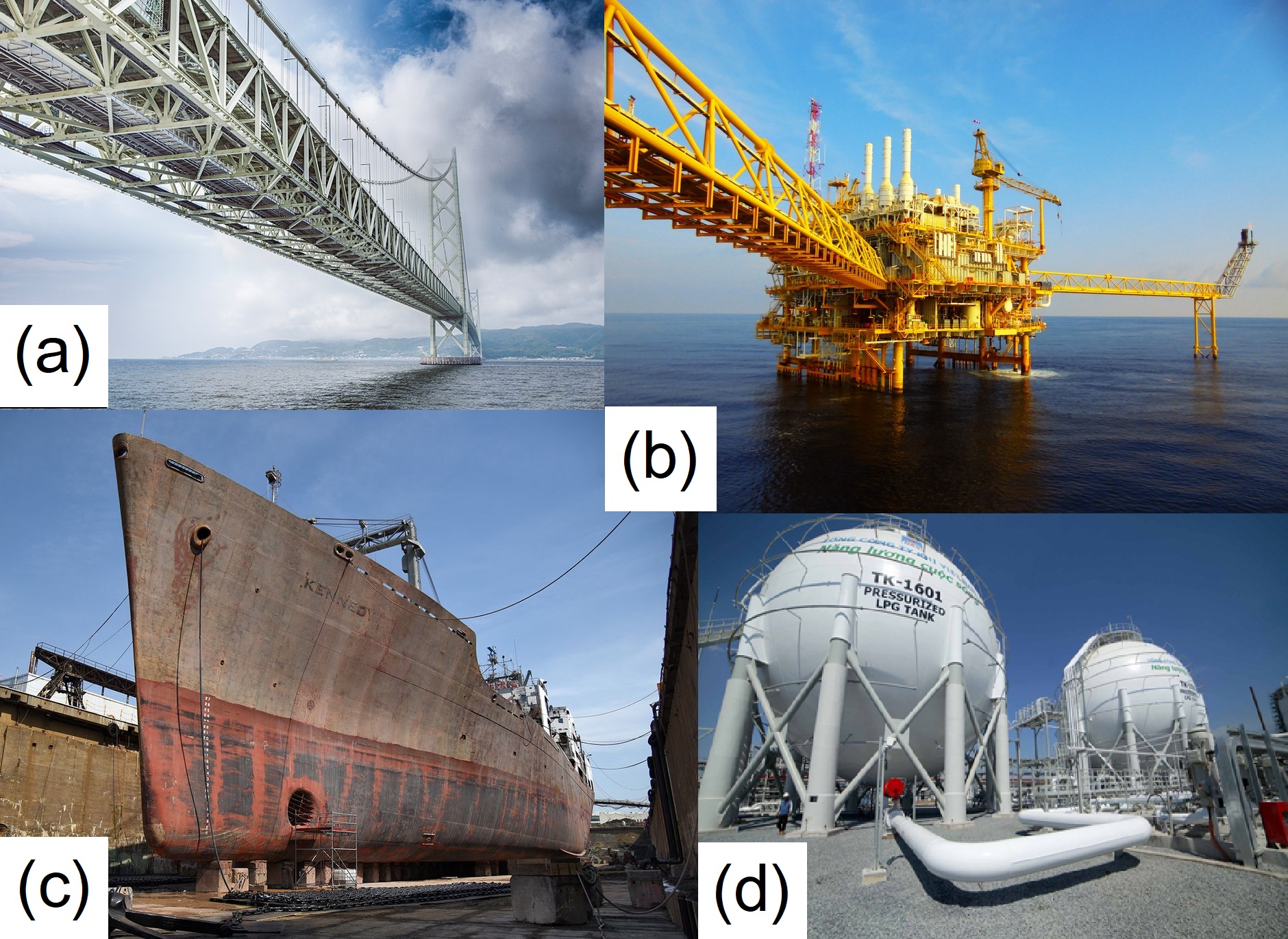}}
    \caption{Typical steel structures: a) Sea-crossing steel bridges. b) Offshore oil rigs. c) Ship shells. d) Oil tanks and pipelines. (Source: Google images)} 
    \label{fig:steelstructures}
\end{figure}

Steel structures are indispensable parts of modern civilization. Typical civil infrastructures including bridges, wind turbines, electric towers, oil rigs, and so on; or vehicles, such as ships and submarines are made of steel. Those structures require maintenance frequently to warrant safety and longevity. Most of these inspections are still conducted manually by professional human inspectors with special devices to inspect visual damages and defects on or inside these structures. This procedure is usually highly time-consuming, costly, and unsafe. For instance, it is dangerous for the inspectors to climb up and hang on cables to inspect remotes area of bridges (Fig. \ref{fig:steelstructures}a), or offshore oil rigs (Fig. \ref{fig:steelstructures}b). Even the inspection of simpler structures such as wind turbines, ship shells (Fig. \ref{fig:steelstructures}c), and gas/oil tanks/piles (Fig. \ref{fig:steelstructures}d) is a big challenge due to their large scales.

To automate the inspection task, using robots carrying sensing tools is an emerging solution. There are several innovative designs, which have been presented in this field in recent years. Several designs were based on conventional wheeled robots \cite{BridgeBot, Zhu_TMECH2012, Guo_ICRA14, Kamdar2015, PL_Allerton2016}, while some made use of tank-like tracks to widen the contacting areas of the robots on steel surfaces \cite{Gu_ICMA2005, NguyenLa_IROS2019, Lee_RAS2013, Seo_TMECH2013}. These approaches can work well on structures with large surfaces such as ships or tanks but face difficulties on complex surfaces frequently encountered on bridges and oil rigs. Some unique developments of climbing robots in this field can be seen in \cite{NguyenLa_SHMII2019, 9196919, Takada_inventions2017}. In this direction, many designs imitated the mobility of climbing animals. A spider-like robot with electromagnets on its feet is reported in \cite{Magnapods}, and a legged robot is developed in \cite{Mazumdar}. An inchworm-like robot \cite{Ward2015ClimbingRF} and a hybrid robot \cite{icra20} are other noted examples. However, the complexity of these robot's mechanics makes controlling them a challenge for real-world applications. Aerial robots provide an alternative for visual inspection tasks without the need for climbing robots. Recent developments make drone control significantly safer in confined spaces \cite{Elios2, Intel_drone2018}. Nonetheless, in-depth inspections of damages inside structures such as analyzing fatigue cracks or steel thickness that requires touched sensors are still challenging for these inspection drones.

\begin{figure*}[!t]
	\centerline{\includegraphics[width=1\linewidth]{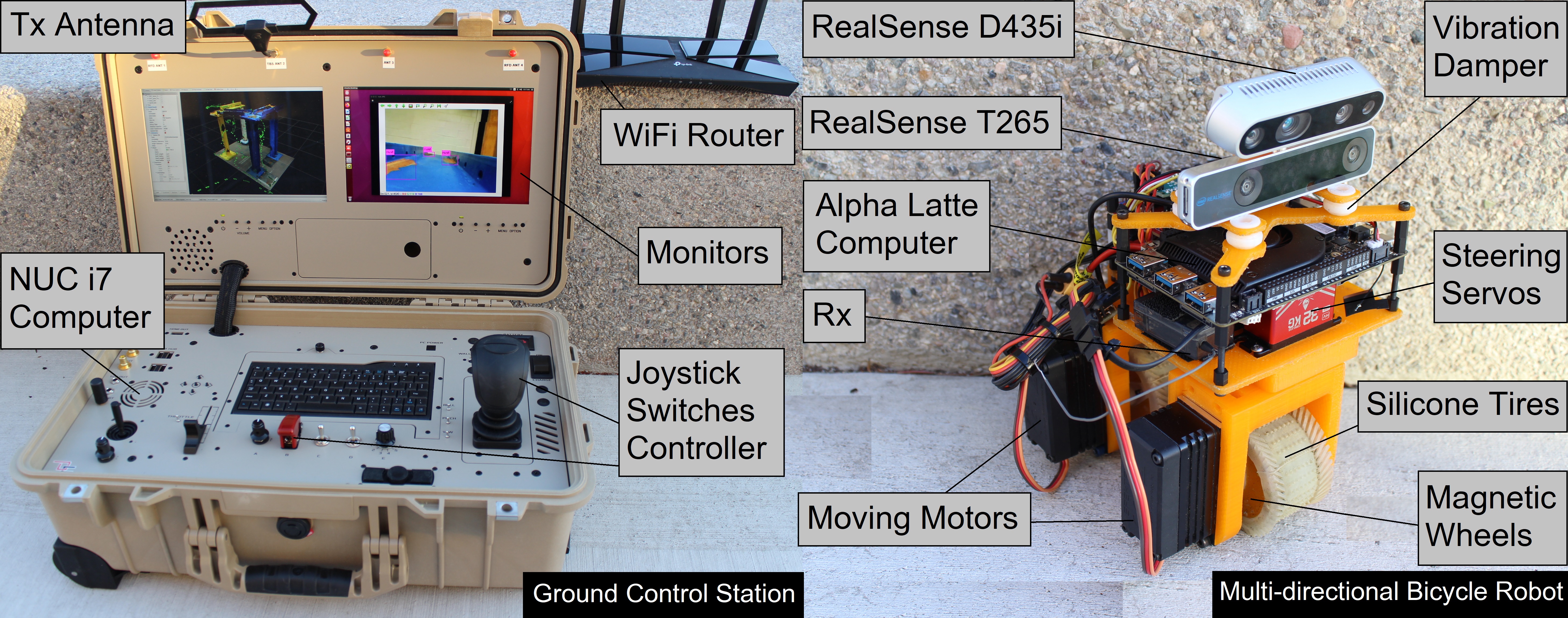}}
	\caption{Multi-directional Bicycle Robot is controlled remotely from a Ground Control Station (GCS) using a joystick via a radio channel. The robot carries a depth camera D435i and a pose tracking camera T265 (both from Intel RealSense). An Alpha Latte computer onboard saves raw data from the sensors and continuously transfers the data back to GCS via a WiFi router. The GCS performs localization, object detection, and visualizes the received data online on its two screens. We use an Intel NUC i7 as the computer for the GCS.}
	\label{fig:system}
	\vspace{-10pt}
\end{figure*}

Interestingly, there were a few bike-like robot designs \cite{Fabien_IJFR2009, Gilles_IJFR2012}. However, the mobility of these robots is still limited in narrow spaces of complex structures, that often require high flexibility of direction changes such as moving sideways. Our design is originated from \cite{Fabien_IJFR2009, Gilles_IJFR2012}, but with an improved optimization on the climbing capability and multi-directional locomotion. Our robot can work on any sorts of ferromagnetic surfaces (flat, curved, or rough), and challenging structures for the previous bike-like robots such as narrow bars or structural transition joints. The proposed robot can generate large adhesive forces by using permanent magnets, which allows our robot to firmly adhere to the steel structures while moving. The multi-directional bicycle design makes the robot flexible on non-standard surfaces of steel structures. The robot is able to overcome difficult obstacles including nuts, bolts, and complex joints between bars or surfaces. Moreover, the robot is equipped with measurement and computational devices including a depth camera, a pose tracking camera, a computer for localization, navigation, and visual inspection. To demonstrate the robot's capabilities and working principles, challenging laboratory and field tests are provided.

\begin{figure}[htbp]
\centerline{\includegraphics[width=0.8\linewidth]{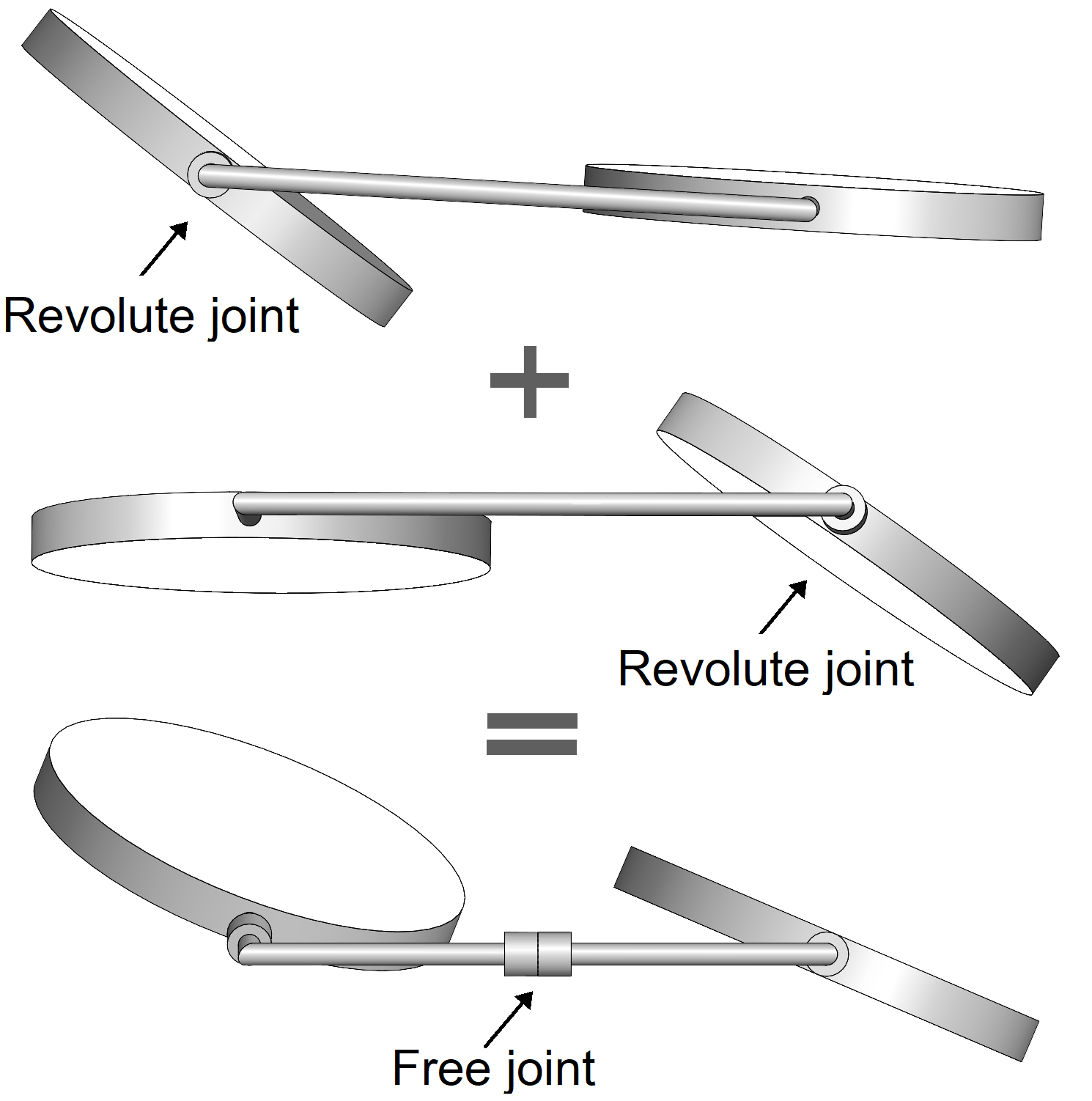}}
    \caption{The above design concept shows the flexibility of our robot. Two independent steering actuators and a free joint help the robot adapt well to challenging working conditions as illustrated in Fig. \ref{fig:perform}. }
    \label{fig:robot}
\vspace{-10pt}
\end{figure}

\section{Overall System}
\label{Sect:overall}

The whole system is depicted in Fig. \ref{fig:system}. On the robot side, we utilize an Intel RealSense D435i camera, which provides both color and depth images. Poses are tracked by using an Intel RealSense T265 tracking camera, which can provide online poses using visual-inertial odometry. We employ a LattePanda Alpha 864 as an onboard computer. This computer is connected with our ground control station (GCS) computer (Intel NUC i7) via a WiFi router. Two computers form a Robot Operating System (ROS \cite{quigley2009ros}) network, in which the GCS acts as the master. The robot is maneuvered remotely by a joystick on the GCS via a radio connection.

Going into the details of our design, given the challenges mentioned earlier for the locomotion systems for reliably inspecting a wide range of surfaces, we consider the following requirements for our robot:
\begin{enumerate}
    
\item  The robot can climb surfaces with a wide range of outer diameters ($\geq$150mm), which are normally encountered on circular tubes or cylindrical surfaces;

\item  The robot can pass thorny, convex, or concave obstacles at structural transition joints on truss structures;

\item The robot can travel on steel structures with complex arrangements of obstacles such as bolts, nuts, and gaps;

\item  The locomotion system can maneuver in narrow areas ($\geq$100mm wide) and can move sideways with large changes in orientation.

\end{enumerate}

\begin{figure*}[!ht]
	\centerline{\includegraphics[width=1\linewidth]{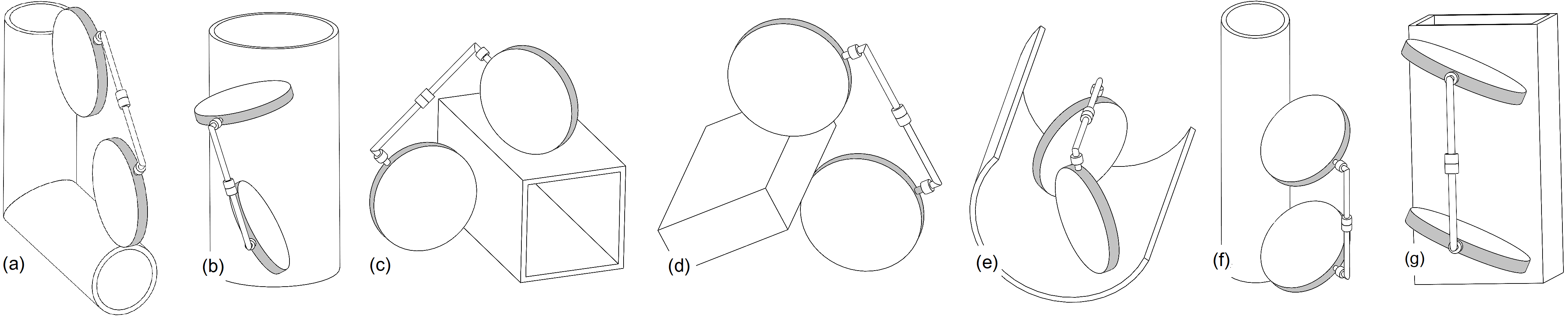}}
	\caption{a) Mode 1 (bicycle-like): The robot can handle cylindrical structures with limited contacting areas. b) Mode 1: The robot changes the direction by first stopping the back wheel. Then, the front steering servo turns 90 degrees, and finally the front wheel moves to help the robot change direction. c) Mode 1: With a free joint, the robot can travel on two intersecting surfaces. d) Mode 1: The robot can traverse on edges that are thicker than the space between its two wheels (4cm). e) Mode 1: The robot is flexible enough travel on the internal surface of a tube. f) Mode 2: Two steering servos turn the wheels at the same angle. The robot moves spirally around a circular tube. In this mode, the robot can also perform well on tube shapes such as rectangles or hexagons. g) Mode 2: The robot can rotate around its body center or move sideways (left, right) at steering angles that are close to 90 degrees.}
	\label{fig:perform}
\end{figure*}

The design concept of our multi-directional bicycle robot is illustrated in Fig. \ref{fig:robot}. Inspired by a bicycle, we use two revolute joints equipping the robot two independent steering actuators, which greatly increase the mobility. An additional free joint in the middle of the robot's body allows its two wheels to fully contact with surfaces of different shapes and sizes. The moving wheels are designed with permanent ring-shaped magnets to generate large adhesive forces. The high mobility of the two steering actuators allows the robot to seamlessly work on two different modes: $(1)$ \textit{Mode 1} (Fig. \ref{fig:perform}a-e): only one steering unit is active, the robot works like a bicycle. The activated steering actuator can be at the front or the back. Working in this mode allows the robot to travel on structures with limited contacting areas (Fig. \ref{fig:perform}a). It can revert the direction immediately without turning, or even turn on spot by stopping the back wheel as illustrated in Fig. \ref{fig:perform}b. In addition, the robot can comfortably climb two intersecting surfaces by virtue of its free joint as shown in Fig. \ref{fig:perform}c. Fig. \ref{fig:perform}d shows that robot can pass edges which are thicker than the space between its two wheels (4cm in our design). Also in this mode, the robot can be flexible enough to travel on an internal surface of a tube (Fig. \ref{fig:perform}e). $(2)$ \textit{Mode 2} (Fig. \ref{fig:perform}f-g): Both the two steering units are active and they are controlled independently and in parallel. The robot can move spirally around a cylinder, move sideways (left, right), or rotate around its center. 

\begin{figure}[ht]
\centerline{\includegraphics[width=0.75\linewidth]{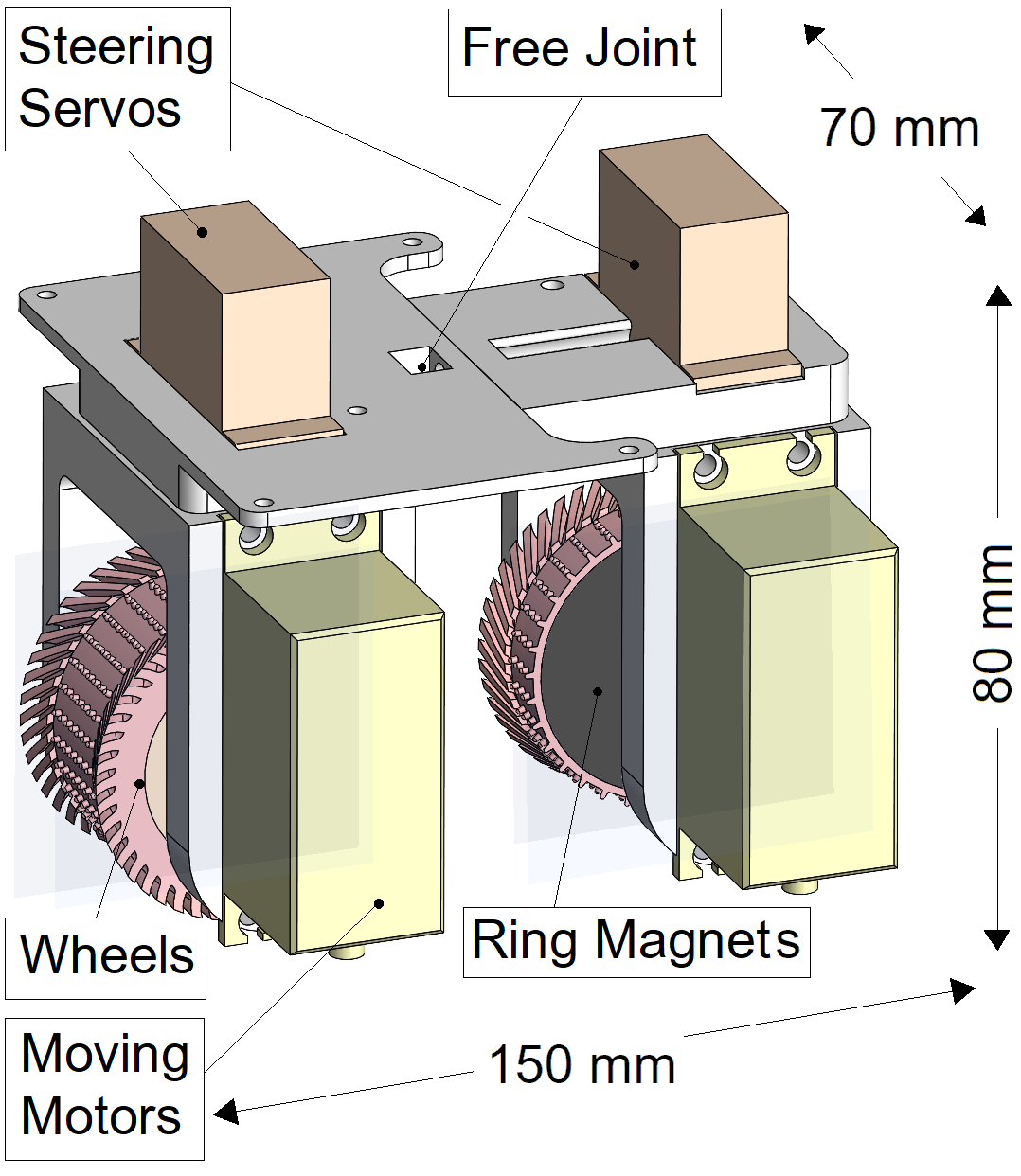}}
    \caption{The 3D mechanical design of our proposed robot.}
    \label{fig:design}
\end{figure}

\section{Mechanical Design and Analysis}
\label{sec:MecDesign}
Fig. \ref{fig:design} shows the overall mechanical design of the bicycle robot. The robot's weight is 1kg (without sensors), while it can carry a 600g of loads (sensors or equipment). The robot is powered by a 3000mAh LiPo battery to allow it to work for up to 1 hour. The robot's dimension is 150mm $\times$ 80mm $\times$ 70mm. The frame is made of plastic for a lightweight robot. The ring magnets are placed at the cores of the wheels, which are covered by silicone tires. The wheels are driven by two high-torque gear DC motors (100kg$\cdot$cm torque each), and the steering actuators are controlled by two servos (32kg$\cdot$cm torque each). The front and the back of the frame are linked by a bearing acting as a free joint. The detail mechanical analysis is presented in the following subsections.

\subsection{Magnetic Wheel Force Analysis}
\label{sec:kinematics}

\begin{figure}[ht]
\centerline{\includegraphics[width=1\linewidth]{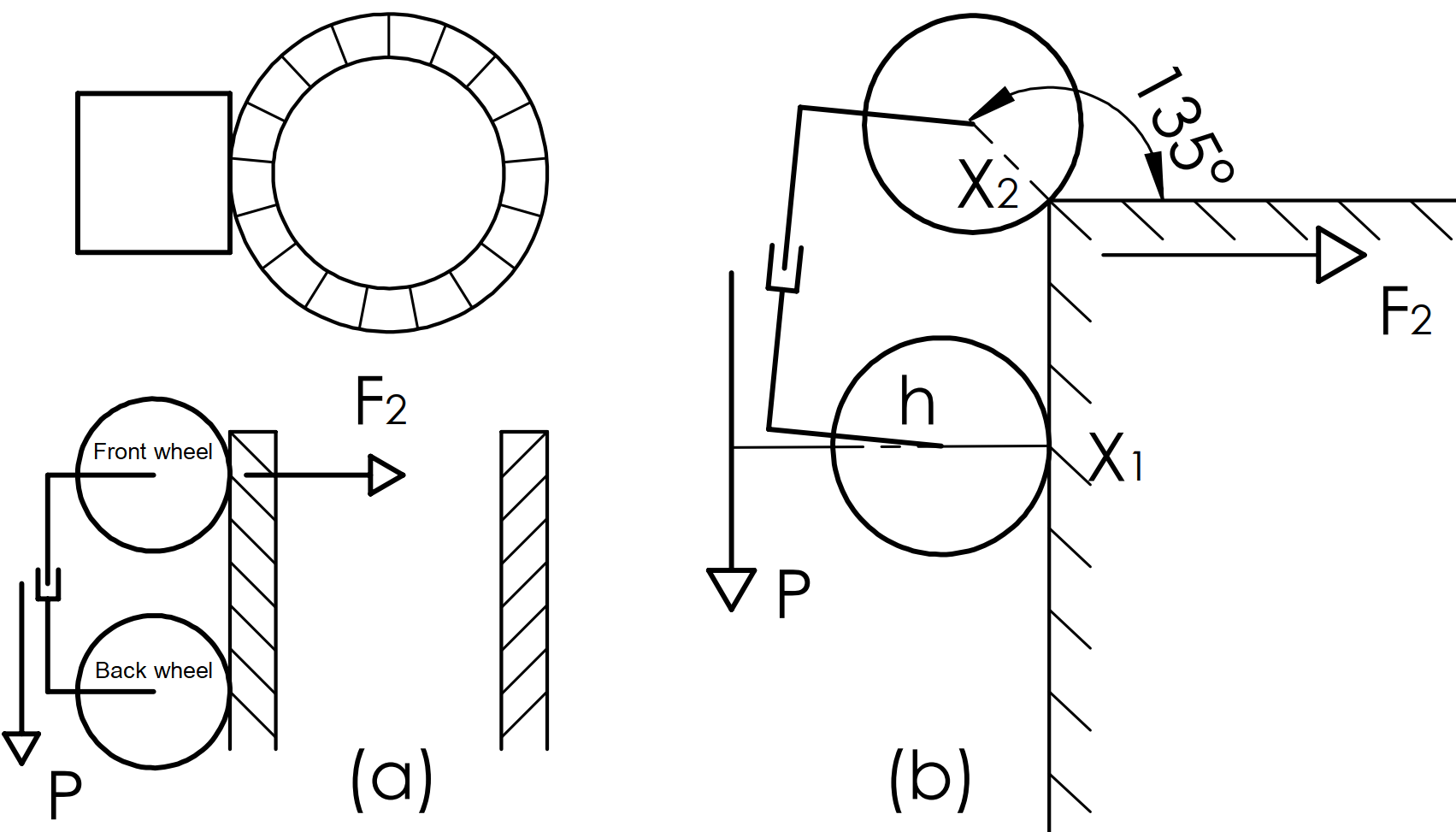}}
    \caption{The situations where the adhesive force is minimal, resulting in a high chance of falling: a) the robot traverses on a thin cylinder (15cm diameter). The contacting area reduces to only a single point for each wheel; b) The adhesive force of the front wheel is significantly reduced when the robot hits an edge.}
    \label{fig:turnover1}
\end{figure}

We analyze the adhesive force that the robot needs to climb reliably in any normally working conditions. We perform the analysis on an extreme situation where the adhesive force between the magnetic wheels and the contacting structures is minimal: the robot climbs a cylindrical tube (Fig. \ref{fig:turnover1}a), and encounters a corner (Fig. \ref{fig:turnover1}b). Here, $X_1$ and $X_2$ are two contacting points of the back wheel and front wheel, respectively. Given $P$ is the robot's weight, $h$ is the distance from the center of mass of the robot to $X_1$. If $F_2$ is an adhesive force of the front wheel at $X_2$ then $F_2$ is at its minimum when the front wheel hits the corner. To keep the robot from falling, the following condition needs to be satisfied:
\begin{equation} \label{e1} 
	F_2 \times X_1X_2 > P \times h \Rightarrow F_2 > \dfrac{Ph}{X_1X_2}.
\end{equation}
According to ISO 3691 \cite{ISO3691} for safe weight lifting, the safety factor of 5 was selected. Therefore, the real adhesive force $F_2$ needs to be at least 5 times greater than the result from the above theoretical calculation (\ref{e1}).

\subsection{Moving Motors Power Analysis}
\begin{figure}[ht]
\centerline{\includegraphics[width=0.7\linewidth]{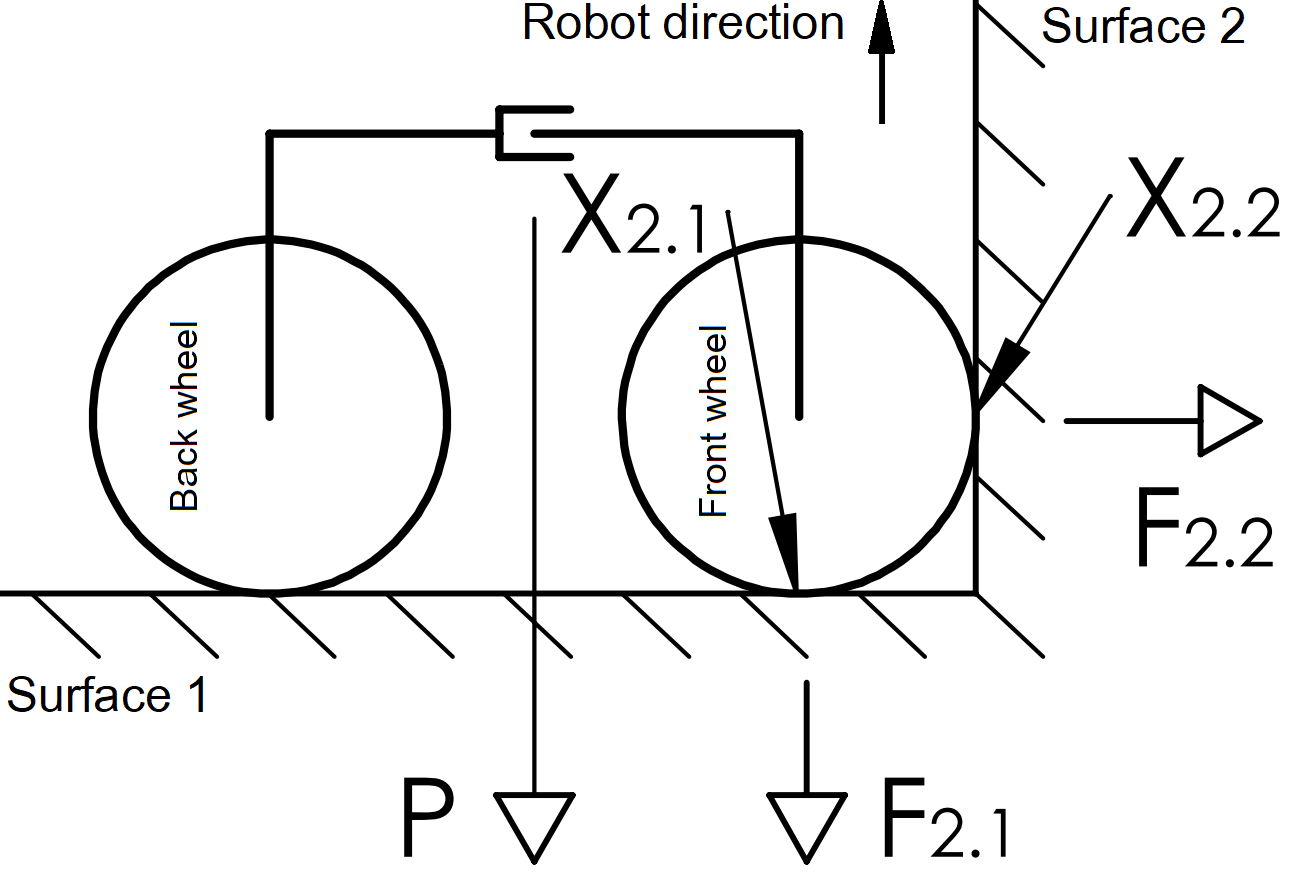}}
    \caption{When the robot passes an internal corner between two surfaces, the robot's load increases significantly.}
    \label{fig:corner}
\end{figure}

This analysis is conducted to calculate the necessary motor torque when the robot stands the highest load. It is when the robot passes an internal corner between two surfaces (Fig. \ref{fig:corner}), the front wheel bears an additional force $F_{2.2}$, which is the adhesive force of the front wheel at surface 2. Similarly, $F_{2.1}$ is the adhesive forces of the front wheel at surface 1. $F_{f2}$ is the friction of the front wheel on surface 2, $r$ is the wheel's radius, $k$ is the static friction coefficient (between silicon and steel in our design). The minimum force of the front wheel that allows the robot to be able to pass the corner must satisfy:
\begin{equation} \label{e3} 
	\dfrac{M_{moving}}{r} > F_{2.1} + F_{f2}.
\end{equation}
Therefore, the moving motor torque needs to satisfy:
\begin{equation} \label{e4} 
	M_{moving} > r\times( F_{2.1} + \dfrac{F_{2.2} + P}{k}).
\end{equation}
According to IEC 60034 \cite{IEC60034}, the actual torque selected to be at least double that of theoretical calculation in (\ref{e4}).

\subsection{Steering Servos Power Analysis}
\begin{figure}[ht]
\centerline{\includegraphics[width=0.6\linewidth]{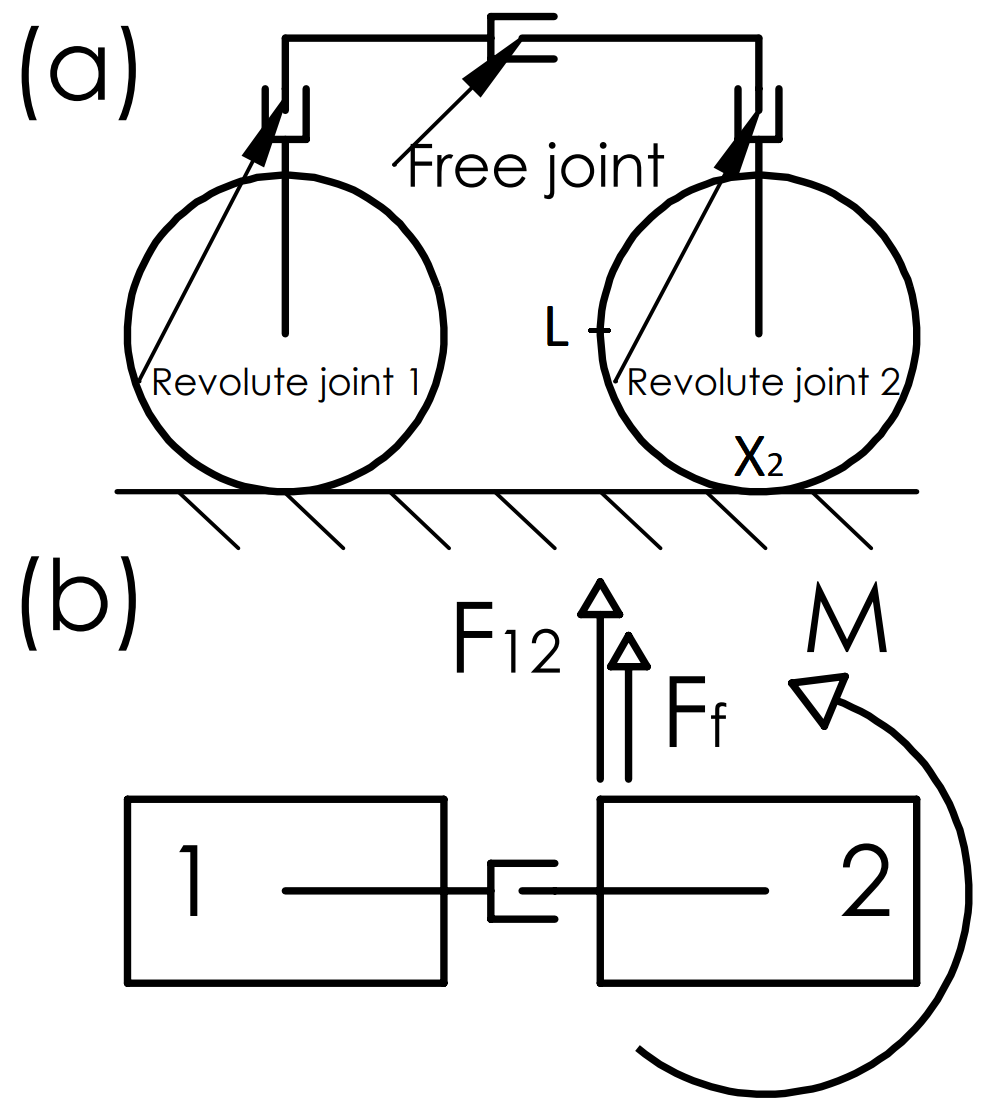}}
    \caption{An experiment is conducted to investigate the load on a steering servo motor. A dynamo-meter is mounted on one wheel's edge (point $L$) to measure the load. The distance from $L$ to the rotating point $X_2$ is $r$ (the wheel's radius). a) Side view. b) Top view. }
    \label{fig:servo}
\end{figure}

An analysis is conducted to investigate the load-torque on the revolute joints. There are two forces as illustrated in Fig. \ref{fig:servo}: the static friction and the attractive force at the two magnetic wheels, with one affects another. Let $F_{12}$ be the adhesive force of wheel 1 affecting wheel 2, $F_{f}$ be the friction at $X_2$. The measured load-force at point $L$ (Fig. \ref{fig:servo}a) has to satisfy the following condition: 
\begin{equation} \label{e5} 
		F_{12} + F_{f}<\dfrac{M_{steering}}{r}.
\end{equation}
Thus, the steering servo torque needs to satisfy:
\begin{equation} \label{e6} 
\Rightarrow
	M_{steering} > r \times ( F_{12} + \dfrac{F_{2} + P}{k}).
\end{equation}
Based on IEC 60034 \cite{IEC60034}, the actual servo's torque is chosen to be at least two-fold compared to that of theoretical calculation in (\ref{e6}).

\subsection{Tire Design}
\begin{figure}[ht]
\centerline{\includegraphics[width=1\linewidth]{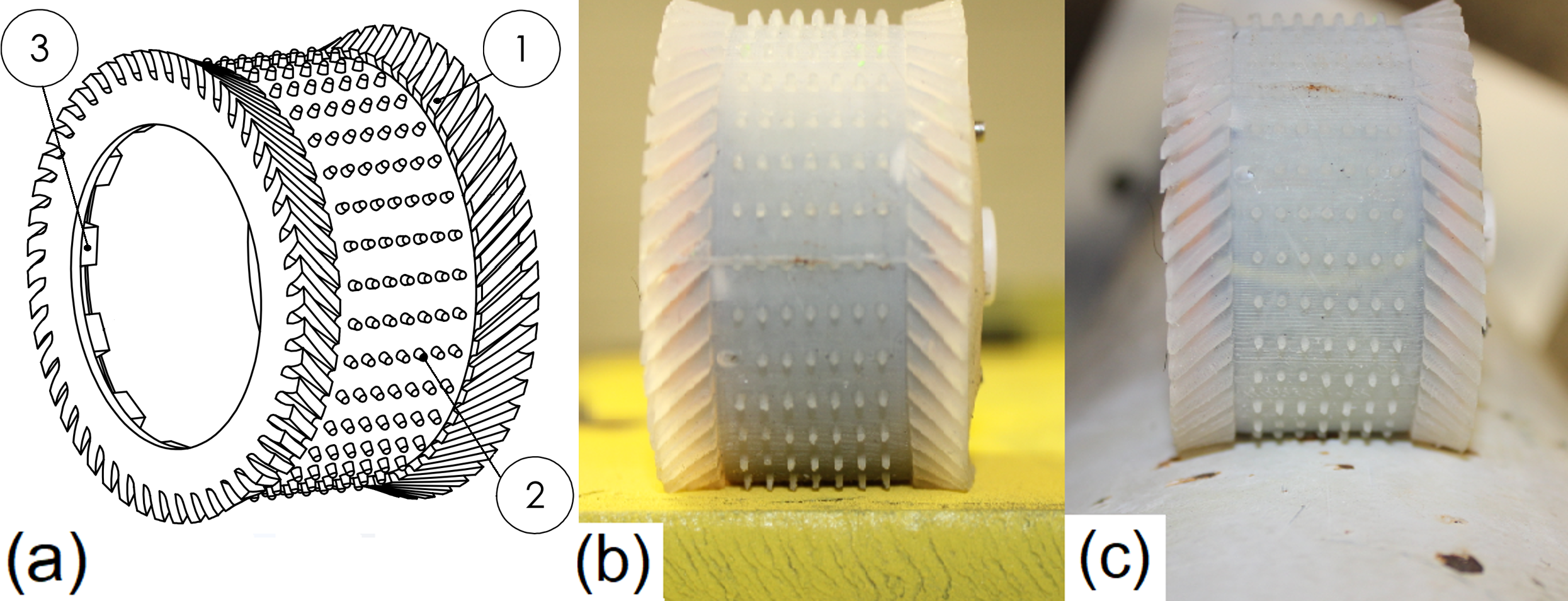}}
    \caption{a) The 3D design of a silicone tire: 1. Spherical-head thorns, 2. Deformable treads, 3. Square keys.  b) The tire adapts to a flat steel surface. c) The tire adapts well 
    on a curved surface of a steel cylinder.}
    \label{fig:tire}
\end{figure}

\begin{figure*}[!ht]
\centerline{\includegraphics[width=\linewidth]{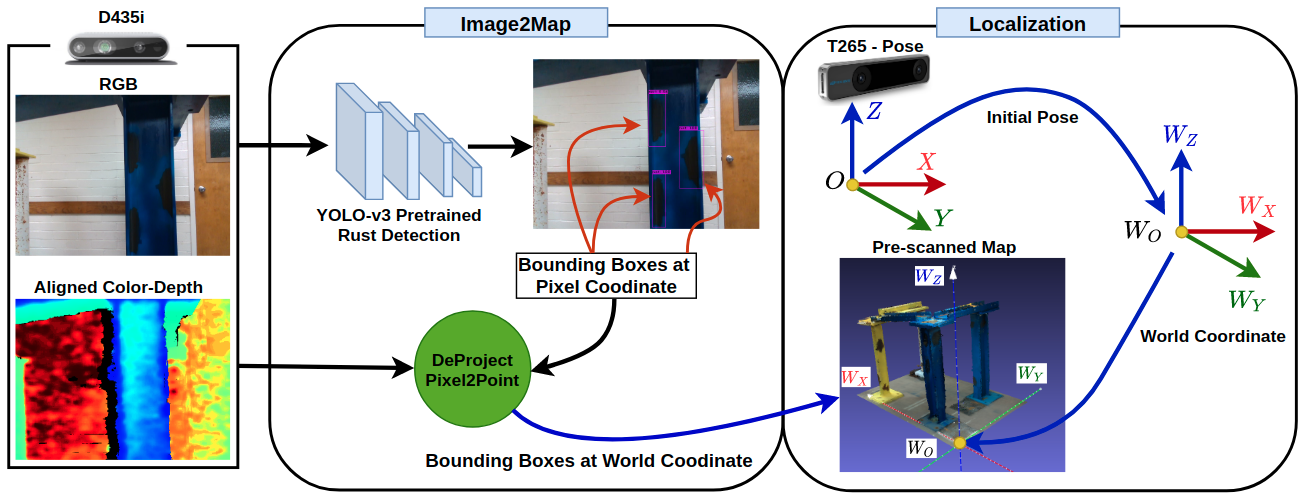}}
    \caption{Descriptions of our software component. Localization is performed by aligning a common world coordinate (which is known to a pre-scanned 3D map of the structure) with the initial pose of T265. We pre-train a YOLO-v3 \cite{redmon2018yolov3} network to detect rusts. The network outputs bounding boxes of detected rusty areas, whose corresponding depth information are retrieved by using aligned color-depth images. Then a pixel2point function will convert these bounding boxes to the world coordinate using the intrinsic parameters of D435i. } 
    \label{fig:software}
\end{figure*}

Specialized tires are designed for our robot to improve the friction between the tires and contacting surfaces. They also help better adapt to various encountered surfaces. The thickness of the tires needs to be small enough to maintain the magnitude of the magnetic force while being not too hard for ease of fabrication. Specifically, the tires are designed to adapt well on flat and cylindrical surfaces with the smallest diameter of 150mm. The proposed design of the tires with deformable curved treads is shown in Fig. \ref{fig:tire}a. We deposited spherical-head thorns and treads for increasing the friction between the tires and the target surfaces and for preventing slipping, especially when the robot operates on a cylindrical surface. Moreover, a dozen of square keys are set to fix and align the tires w.r.t. the inside cores. Due to the deformation of the tire material and trenches, these treads can deform and arrange tidily when traveling on a planar surface, promising a flat surface contacting (Fig. \ref{fig:tire}b). When the robot travels on curved surfaces, curved treads will also fit well to the target surfaces as shown in Fig. \ref{fig:tire}c. Regarding the material, it needs to be soft with low viscosity, in order to fill in the mould cavities and to avoid air bubbles. We chose DragonSkin 30 (Smooth On Inc., USA) silicone, which fits our technical requirements.

\section{Experimental Evaluation}
In this section, we demonstrate an application of our robot to detect rusts on steel bridges. This showcase is general and it can be applied to other steel structure inspection. We first describe the software structure, followed by laboratory and field tests. The hardware is previously described in Section \ref{Sect:overall}.

\subsection{Software Structure}
The software structure is described in Fig. \ref{fig:software}. While the onboard computer collects the depth and color images, these images are transferred wirelessly back to the GCS, where localization and rust detection are performed. Unlike a similar system in \cite{dang2020autonomous}, where all computation was performed onboard, our structure does not require onboard powerful computers and computationally expensive resources, which results in a reduction of the robot's load.
\par Localization of the robot is performed w.r.t. a common (world) coordinate known to both T265's pose coordinate system and a pre-scanned map. The world coordinate is the first pose of T265 and it is manually chosen in the pre-scanned map. Ideally, the mapping should also be done by the robot and it seems that our setting is sub-optimal. However, we found that for a single robot, the onboard camera cannot avoid blind spots due to its limited view, especially at abruptly turning points. Hence, it is challenging to merge neighboring point clouds to create a useful map. To overcome this issue, future work may consider using a team of robots or actively controlling the onboard camera. Rust detection on RGB images is performed by a pre-trained YOLO-v3 \cite{redmon2018yolov3} network. We use YOLO-ROS \cite{bjelonicYolo2018} to use the trained network in ROS. While each resulting bounding box $[b_x, b_y]$ of detected rusty areas is on the pixel coordinate, we want to localize them on the map. We do that by comparing its coordinate with the corresponding pixel on an aligned color-depth image to retrieve the corresponding depth information $b_z$. Next, $[b_x, b_y, b_z]$ is fed to a deprojection function to convert to the world coordinate using the intrinsic parameters of D435i. 

\begin{figure}[htbp]
\centerline{\includegraphics[width=1\linewidth]{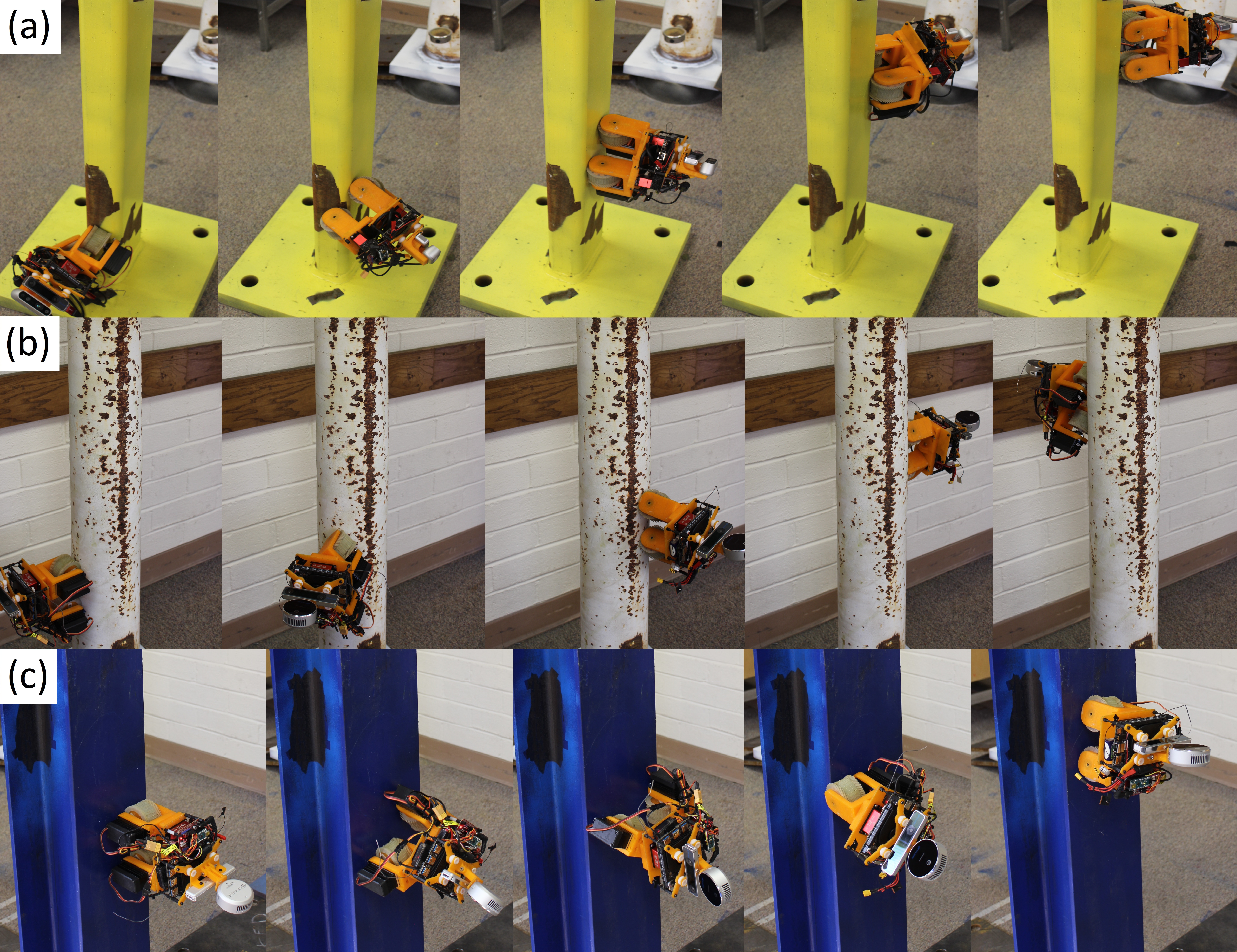}}
    \caption{ a) A test of moving around thin rectangular tube. Robot has to combine mode 1 and 2 for a smooth travel. b) A test of moving sideways in mode 2. The robot travels spirally outside a cylindrical structure. c) A turning test in mode 1. The robot's head is turned 180 degrees on a narrow surface. The movement is depicted from left to right.}
    \label{fig:climbing1}
\end{figure}

\begin{figure}[htbp]
\centerline{\includegraphics[width=1\linewidth]{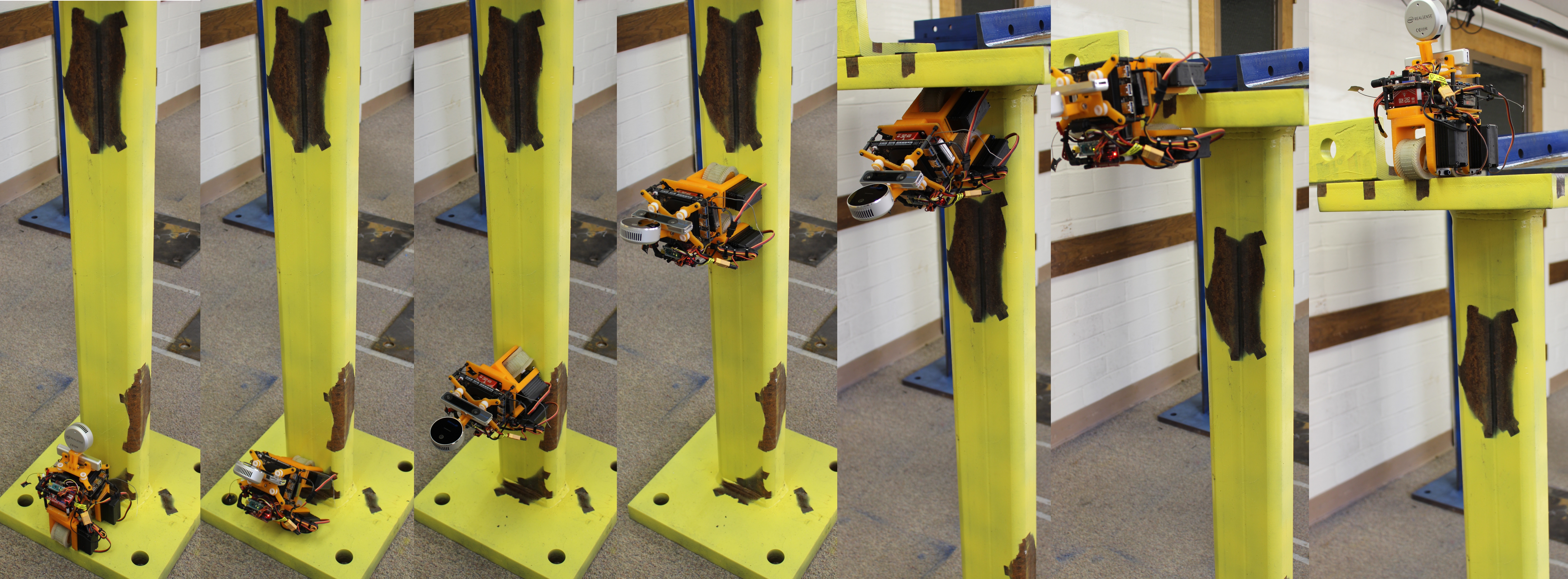}}
    \caption{Passing corners test for mode 1. The robot traverses 90 degrees on internal and external corners. The movement is depicted from left to right.}
    \label{fig:indoor1}
\end{figure}

\subsection{Laboratory Tests}
An indoor structure comprised of typical parts of general steel structures (cylinder, L, I, U shaped beams) with structural transition joints is built to validate the robot's locomotion. Our robot can traverse smoothly to any location in the testing structure. In mode 1, the robot can handle well most testing situations. However, mode 2 is necessary when operating in narrow spaces. For instance, as Fig. \ref{fig:climbing1}a shown, the robots need to combine mode 1 and 2 for travelling smoothly on a rectangular tube. Fig. \ref{fig:climbing1}b illustrates mode 2 while the robot working on a cylindrical shape. Fig. \ref{fig:climbing1}c illustrates how the robot changes orientation in mode 1 on an I-shaped beam. Fig. \ref{fig:indoor1} illustrates when the robot crosses convex and concave obstacles of a rectangular tube.


\begin{figure}[ht]
\centerline{\includegraphics[width=1\linewidth]{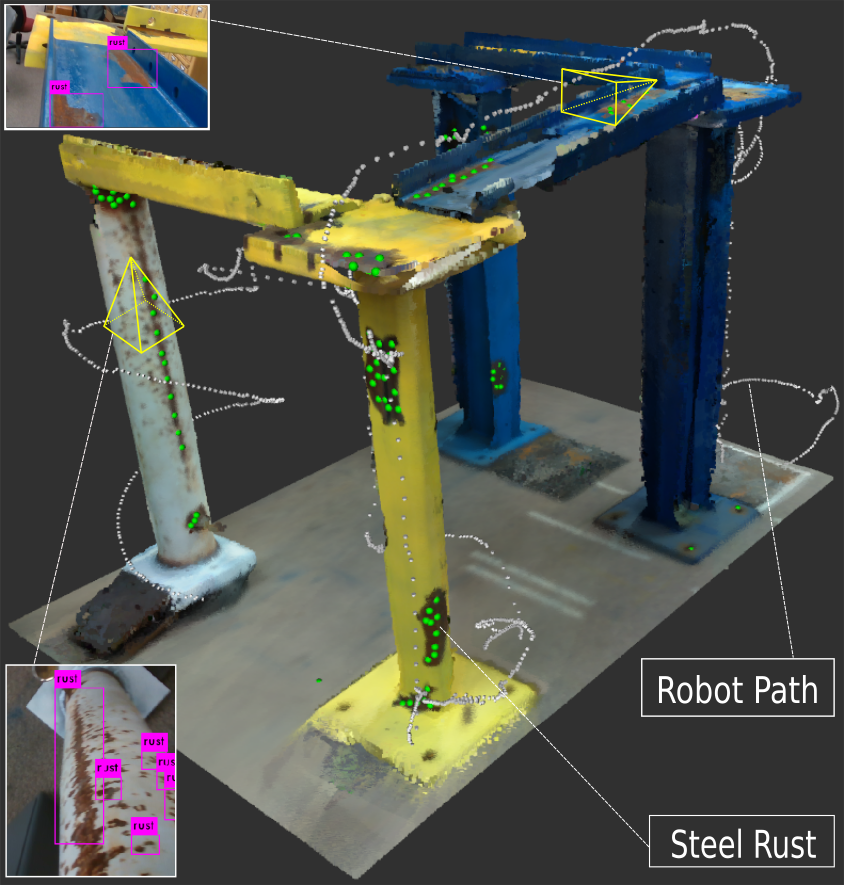}}
    \caption{The figure shows the inspection result for the indoor steel bridge. The 3D mapped positions of rusts are indicated by green spheres on the bridge. An example of the robot paths is indicated by a white trace. Two viewpoints from the robot are shown with detected rusts marked by pink boxes.}
    \label{fig:indoor-bridge}
\end{figure}

\subsection{Field Tests}
\begin{figure}[ht]
\centerline{\includegraphics[width=1\linewidth]{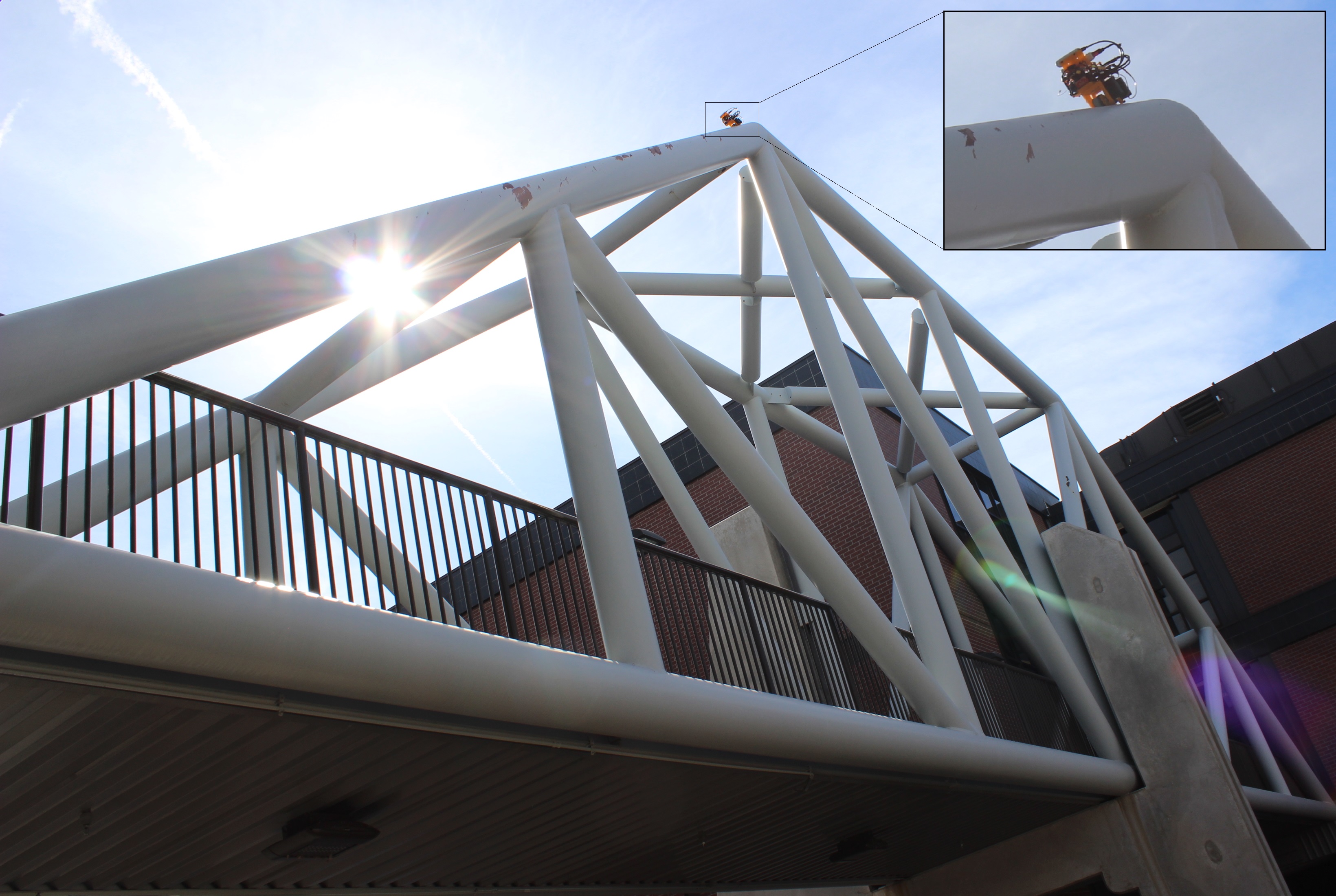}}
    \caption{Robot operating on a cylindrical steel bridge.}
    \label{fig:outdoor-bridge}
\end{figure}

We deploy our robot to perform a structural inspection for one indoor and one outdoor steel bridge. We first scan the indoor bridge manually using Dot3D\footnote{https://www.dotproduct3d.com/dot3dpro.html} software. During experimenting with other approaches, we find this software is easier to use and can produce good 3D maps, which have better qualities than the ones produced by open-source approaches such as RTAB-Map \cite{labbe2019rtab} or VoxBlox \cite{oleynikova2017voxblox}. The inspection result of the indoor bridge is shown in Fig. \ref{fig:indoor-bridge}. In the figure, we can observe that rusty areas, which are marked by collective green spheres, are correctly identified and localized. We also show an example of the robot paths, illustrated by a white trace. Additional color images from the onboard D435i camera are shown with detected rusts marked by pink boxes. For the outdoor bridge, we only test the locomotion capability of our robot due to the difficulty of scanning this much bigger structure. Fig. \ref{fig:outdoor-bridge} shows our robot operating on surfaces of this bridge. Regarding the online performance, please refer to our supplementary video at \url{https://youtu.be/g35J7auf6oM}.

\label{sec:Result}
\section{Conclusions and Future Work}
\label{sec:Conclusion}

This paper presents a novel development of a multi-directional bicycle robot, which is capable of agilely climbing different steel structures to perform the structural inspection. The robot design is implemented and validated on laboratory testing structures and a cylindrical steel bridge. During all tests, the robot is able to firmly adhere to steel structures with various challenging levels. A rigorous theoretical analysis of the magnetic adhesive forces has been performed to confirm that the robot is able to work reliably in challenging situations of real applications. The robot can also operate well in limited spaces. A demonstration of deploying our robot to perform rust detection on steel bridges is shown. The visual data is collected and transferred to a ground station for visualization and processing to create a 3D map of the structures with marked locations of rusts. In the future, the robot can be further equipped with non-destructive testing (NDT) sensing modules for more in-depth inspections (steel thickness, fatigue cracks, and so on).

From the autonomy perspective, future directions can consider performing autonomous map construction, localization, navigation, and object detection all onboard. Given the known maps, it is also possible to perform path planning for autonomous inspections or even learning-based methods such as reinforcement learning to train the robot to carry the inspection task more efficiently.

\bibliographystyle{IEEEtran}
\balance
\bibliography{RefFile}

\end{document}